\documentclass[10pt,twocolumn,letterpaper]{article}
\PassOptionsToPackage{table}{xcolor}
\usepackage[pagenumbers]{wacv} 
\usepackage{svg}
\usepackage{amsfonts}
\usepackage{amsmath}
\usepackage{amssymb}
\usepackage{mathtools}
\usepackage{array}
\usepackage{bm}
\usepackage{enumitem}
\usepackage{graphicx}
\usepackage{microtype}
\usepackage{nicefrac}
\usepackage{placeins}
\usepackage[table]{xcolor}
\usepackage{pifont}
\usepackage{url}
\usepackage{xcolor}
\usepackage{xspace}
\usepackage[accsupp]{axessibility}  
\usepackage{booktabs}
\usepackage{multirow}
\usepackage{rotating}
\usepackage{adjustbox}
\usepackage{wrapfig}
\usepackage{tcolorbox}
\usepackage{makecell}
\usepackage{xcolor}   
\usepackage{siunitx}
\definecolor{Gray}{gray}{0.2}
\definecolor{lightgray}{gray}{0.92}
\definecolor{LightCyan}{rgb}{0.88,0.95,1}
\definecolor{OurColor}{rgb}{0.855, 0.937, 0.957}
\definecolor{customgray}{gray}{0.35}
\definecolor{bandgray}{gray}{0.85}

\definecolor{wacvblue}{rgb}{0.21,0.49,0.74}
\usepackage[pagebackref,breaklinks,colorlinks,allcolors=wacvblue]{hyperref}

\newcolumntype{C}{S[
  table-format = -1.3,
  table-number-alignment = center,
  table-fixed-width = true,
  table-column-width = 2.2em
]}
\newcolumntype{A}{S[
  table-format = 1.2,
  table-number-alignment = center,
  table-fixed-width = true,
  table-column-width = 2.2em
]}

\newlength{\qbcell}
\newlength{\qbwidth}

\definecolor{qbstrip}{gray}{0.93}
\definecolor{qbkey}{rgb}{0.54,0.35,0.17}

\definecolor{qbsOne}{HTML}{7F1D1D}
\definecolor{qbsTwo}{HTML}{A9491A}
\definecolor{qbsThree}{HTML}{7D6A12}
\definecolor{qbsFour}{HTML}{47732C}
\definecolor{qbsFive}{HTML}{14532D}
\definecolor{qbsZero}{HTML}{4A4A4A}
\definecolor{qbsNaN}{HTML}{E6E4E0}
\newlength{\qbchipw}
\newcommand{\qbchip}[3]{\colorbox{#1}{\makebox[\qbchipw][c]{%
  \textcolor{#2}{\fontsize{6}{7}\selectfont\bfseries\strut #3}}}}
\newcommand{\qbscorex}[2]{%
  \ifcsname qbsc@#1\endcsname
    \def\qbtmp{\csname qbsc@#1\endcsname}%
  \else
    \def\qbtmp{\qbchip{qbsNaN}{black}}%
  \fi
  \qbtmp{#2}}

\expandafter\def\csname qbsc@0\endcsname{\qbchip{qbsZero}{white}}
\expandafter\def\csname qbsc@1\endcsname{\qbchip{qbsOne}{white}}
\expandafter\def\csname qbsc@2\endcsname{\qbchip{qbsTwo}{white}}
\expandafter\def\csname qbsc@3\endcsname{\qbchip{qbsThree}{white}}
\expandafter\def\csname qbsc@4\endcsname{\qbchip{qbsFour}{white}}
\expandafter\def\csname qbsc@5\endcsname{\qbchip{qbsFive}{white}}

\def\confName{WACV}
\def\confYear{2027}

\title{What Does Animal Re-Identification Learn?\\
Linear Biological Concepts and Their Origins in Visual Representations}

\author{
Robert Nolting$^{1,2,3,*}$,
Alexandra Schild$^{1,2,*}$,
Moritz Weckbecker$^{3}$,
Maximilian Schall$^{1,2}$, \\
Gerard de Melo$^{1,2}$ \vspace*{1.5mm}\\
$^{1}$Hasso-Plattner Institute, Germany $\qquad$
$^{2}$University of Potsdam, Germany \\
$^{3}$Department of Artificial Intelligence, Fraunhofer Heinrich Hertz Institute, Germany \\
$^{*}$Equal contribution\vspace*{0.5mm}\\
{\tt\small\{robert.nolting\}@student.hpi.de} \\
{\tt\small\{aleksandra.kudaeva\}@hpi.de} \\
}

\begin{document}
\maketitle
\begin{abstract} 
Conservation increasingly relies on camera traps that collect more wildlife imagery than experts can manually analyze, making animal re-identification (Re-ID) essential for monitoring individuals and populations. Yet understanding which cues drive model decisions is challenging for ViT-based Re-ID models, whose metric-learning objectives provide no explicit supervision for biological concepts. We ask whether such models nonetheless organize their representations along biologically meaningful axes. Using a DINOv3 backbone fine-tuned for Western lowland gorilla Re-ID with triplet-margin loss, we find that sex and age emerge as linear directions that generalize to held-out individuals, reaching up to 0.91 AUROC and being recoverable from a single image per individual. Activation steering further shows that the sex direction is causally used by the model, flipping a significant fraction of predictions to the opposite sex. Comparing off-the-shelf and fine-tuned backbones shows that Re-ID training does not create these concepts, but relocates them across the network. Finally, data attribution reveals that the representation we find reflects a graded biological axis, is redundantly encoded across the population and shaped by visually ambiguous individuals. Together, these findings show how interpretability can uncover both the biological structure and failure modes of Re-ID representations, providing a step toward auditable computer vision for wildlife monitoring. Our code is available: \url{https://alexandraschild.github.io/bio-concepts/}.
\end{abstract}

\vspace*{-2mm}
\section{Introduction}
\label{sec:intro}

Conservation efforts increasingly rely on motion-triggered camera traps, which collect far more observations of animals in the wild than can be manually annotated and analyzed \cite{norouzzadeh2018automatically}. Animal re-identification (Re-ID) can help turn these large-scale data collections into valuable information about individual animals and populations. However, for animal conservation efforts, the trustworthiness of model predictions matters alongside their accuracy: practitioners need to understand why a model identifies an individual and whether its decisions rely on biologically meaningful cues or spurious correlations. This is particularly important for modern deep learning-based Re-ID models, whose learned representations are difficult to interpret. Interpretability techniques can reveal which cues drive identification and whether the model relies on misleading ones, which is essential for evaluating whether Re-ID models can be trusted in long-term, real-world conservation monitoring.

One promising route toward understanding model behavior is to identify human-meaningful concepts encoded by it. A growing body of work suggests that many semantic attributes are represented as approximately linear directions in neural activation spaces \cite{park2023linear}. Such directions can be identified with linear probes, validated against external ground truth, and, importantly, causally manipulated through activation interventions. This perspective turns an otherwise opaque activation into a representation that can be inspected and experimentally tested.

However, existing evidence for linear concept representations comes predominantly from models trained with explicit class supervision or generative objectives. It remains unclear whether comparable interpretable structures emerge under metric learning, the objective commonly used for Re-ID and one that does not require a classification head. This raises a fundamental question: \emph{can a model trained only to distinguish individual animals nevertheless organize its internal representation along biologically meaningful axes such as sex and age?}

\begin{figure*}[t]
\centering
\includegraphics[width=\textwidth]{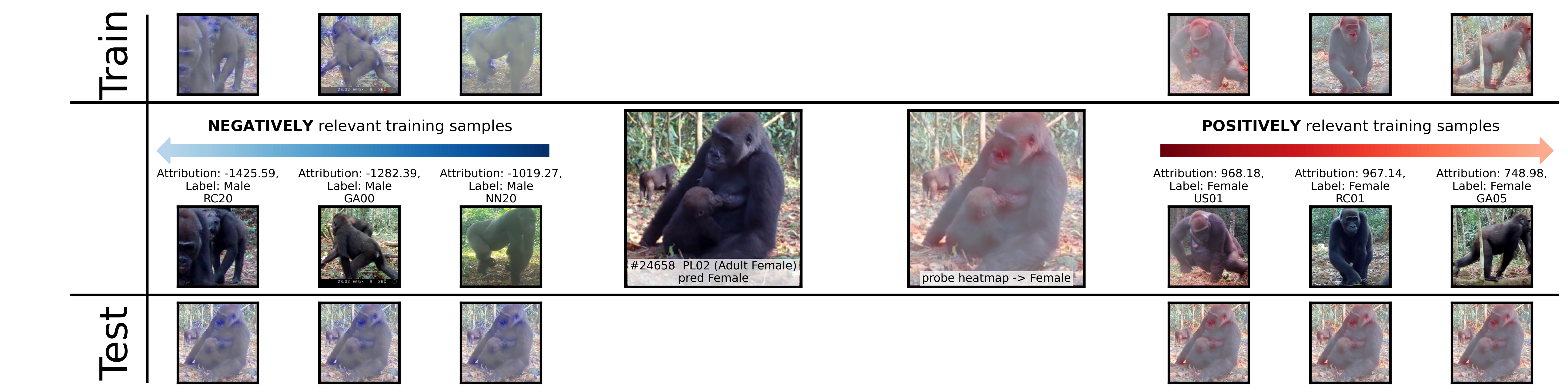}
\caption{Feature attribution for a female test image, produced by DualXDA. The middle image shows a female gorilla that was correctly predicted by our activation probe, together with an LRP heatmap showing which parts of the image were important for the classification. The right side shows the training images that pulled the prediction most towards female, the left side towards male. The heatmaps show that relevance concentrates on facial and anatomical regions but also on the nursing infant, suggesting a potential shortcut.}
\label{fig:dualxda_female}
\end{figure*}

We investigate this question in the context of Western lowland gorilla re-identification, using a DINOv3-based model trained to distinguish individual gorillas \cite{simeoni2025dinov3}. Its backbone is fine-tuned on labelled images of individual gorillas using a triplet-margin loss. 
Rather than treating interpretability as a purely post-hoc visualization problem, we study the resulting representation from several complementary perspectives. First, we ask whether biologically meaningful concepts are linearly encoded across the model's layers and whether these directions generalize to individuals not seen during training. Second, we assess whether the resulting directions 
are actually used by the model, using causal interventions during the forward pass. Third, we compare the geometry of the stock DINOv3 backbone with that of its Re-ID-fine-tuned counterpart to determine how task-specific training alters existing concepts. Finally, we rely on data attribution to examine what the learned representation is built from: which training individuals contribute to a concept direction, how influence is distributed across the training set, and how robust the representation is to removing individual examples.

Our analysis reveals a consistent picture: biologically meaningful concepts emerge as structured, causal, and highly redundant components of the Re-ID representation, with Re-ID fine-tuning substantially reshaping where these concepts reside. Moreover, the learned representation reflects the structure of the underlying biology: concept directions form graded axes rather than discrete categories, and difficult biological cases disproportionately shape the resulting representation. At the same time, attribution exposes both useful diagnostics and potentially misleading shortcuts, highlighting why understanding a deployed Re-ID system requires going beyond predictive accuracy alone.

In summary, our findings are:

\begin{itemize}

\item \textbf{Linear concepts under metric learning.}
A ViT trained solely for gorilla Re-ID encodes sex and age as approximately linear directions, reaching up to 0.91 AUROC on held-out individuals and remaining detectable from a single image per individual.

\item \textbf{Functional role in retrieval.}
Activation steering along the sex direction shifts identity predictions toward individuals of the opposite sex, while off-manifold diagnostics distinguish this targeted effect from representation collapse.

\item \textbf{Fine-tuning reshapes existing concepts.}
Pretrained and Re-ID-fine-tuned DINOv3 exhibit similar concept separability, but their directions increasingly diverge across layers, with sex reaching only $\sim$0.25 cosine similarity in the final layer. Fine-tuning therefore reorganizes rather than creates the concept representation.

\item \textbf{Biologically structured and redundant representation.}
The learned sex direction forms a graded axis reflecting sexual dimorphism across development. Visually ambiguous blackbacks are disproportionately represented among its constituting examples, yet removing any individual has little effect on held-out performance.

\item \textbf{Attribution reveals potential shortcuts.}
Feature attribution surfaces a biologically plausible but potentially spurious shortcut: on images that show female gorillas nursing an infant, heatmaps show attribution placed on the infant.
\end{itemize}

\section{Related Work}
\label{sec:related}

\subsection{Animal Re-Identification}

Animal Re-ID seeks to match individuals across images or video under challenging conditions such as pose variation, occlusion, and limited labeled data. Early methods relied on hand-crafted individual-specific patterns, but the field has increasingly adopted deep metric learning with learned embeddings \cite{hermans2017defense,bergamini2018multi,chen2020panda,clapham2020bear}. More recently, Vision Transformers and self-supervised foundation models such as DINOv2 and DINOv3 have enabled more robust and transferable representations, reducing reliance on large species-specific training sets \cite{dosovitskiy2020image,he2021transreid,oquab2023dinov2,simeoni2025dinov3}.

These developments have also advanced primate and gorilla Re-ID, with recent systems enabling identification of wild gorillas at scale \cite{laskowski2023gorillavision,schall2026gorillawatch}. This makes gorilla Re-ID a useful setting for studying not only identification performance, but also the biological information encoded in modern visual representations.

While using deep learning models made Re-ID more accurate and robust, what determines model decisions in each instance and whether the model is using sensible visual cues is still an open question. Some recent studies analyze attention patterns to understand if a Re-ID model uses meaningful visual cues and confirm that existing models often base their predictions on spurious correlations \cite{rueda2026we, schall2026gorillawatch}. 


However, to our knowledge, there is no existing work exploring if biological concepts are identifiable in the vision representations and if they guide re-identification.

\subsection{Linear Concept Representations}

Neural networks encode information in distributed activation spaces, from which concepts can be recovered using probing methods \cite{hinton1986learning,alain2016understanding}. A growing body of work shows that high-level concepts can often be represented as approximately linear directions, including semantic attributes in language \cite{mikolov2013efficientestimationwordrepresentations,marks2023geometry,gurnee2024language} and human-defined concepts in vision \cite{kim2018interpretability}. These directions can also be causally tested: interventions along semantic directions can modify model behavior or generated attributes \cite{shen2020interpreting,arditi2024refusal}. Such findings motivate the Linear Representation Hypothesis, which proposes that high-level concepts are encoded as linear directions in neural activation spaces \cite{park2023linear}. However, linear probes alone do not establish that a concept is used by the downstream task \cite{ravichander2021probing}. Causal intervention therefore provides a stronger test of task relevance.

Most prior work on linear representations studies supervised classifiers or generative models, where the training objective directly encourages semantic structure. Re-ID embedding models represent a different regime: they have no classification head and are trained with a metric-learning objective that only constrains distances between individual identities. Whether biologically meaningful concepts emerge as linear directions under this objective, and whether these directions causally influence identity predictions, remains largely unexplored. We address this gap through layer-wise probing and activation steering.

\subsection{Data Attribution}

Data attribution (DA) asks which training examples are responsible for a model's predictions or learned behavior. The canonical formulation is counterfactual: it measures how a model's prediction changes when a training example is removed and the model retrained \cite{koh2017understanding}. This notion is usually approximated due to the high computational cost \cite{park2023trak}.

A second family of methods characterizes a prediction through decomposition. Representer-point methods express a model's output as an additive sum of contributions from individual training examples \cite{yeh2018representer}.
DualXDA \cite{yolcu2025sparseefficientexplainabledata} is a representer-point method to perform efficient DA via an SVM surrogate model. 

Closest to our setting, Konz \emph{et al.}~\cite{konz2023attributing} attribute linear concept probes on a ResNet-18 back to their ImageNet training data, identifying the images responsible for the probe learning a concept like 'Snakes'. We pursue a similar goal but apply a different DA method to the activations of a ViT trained specifically for gorilla Re-ID.

\section{Method}
\label{sec:method}

\subsection{Residual-Stream Representation}
\label{sec:resstream_representation}

To extract activations from our Vision Transformer when given an input image, we extract the residual-stream activations $\mathbf{X}^{(l)} \in \mathbb{R}^{T\times d}$ at each transformer layer $l$, where $T$ is the number of tokens and $d$ the hidden layer dimensionality. We obtain an image-level representation by averaging the token activations:

\begin{equation}
\mathbf{x}^{(l)} =
\frac{1}{T}\sum_{t=1}^{T}\mathbf{X}^{(l)}_t.
\end{equation}

For the Re-ID model, the final representation is further projected into the embedding space used for identity retrieval. We study both the pretrained DINOv3 backbone and the same backbone after Re-ID fine-tuning, allowing us to measure how the metric-learning objective changes the geometry of existing concepts.

\subsection{Linear Concept Probing}
\label{sec:probing}

Following work on linear concept representations \cite{park2023linear}, we use linear probes to test whether biological concepts are encoded as directions in the Re-ID representation. Unlike prior applications focused primarily on supervised or generative models, we study their emergence under metric learning, where the model is trained only to distinguish individual identities and receives no concept supervision.

Given activations $\mathbf{x}^{(l)}$ and binary concept labels $y\in{0,1}$, we learn a direction $\mathbf{t}^{(l)}$ using either $\ell_2$-regularized logistic regression or a linear SVM. We additionally evaluate the parameter-free difference-in-means (DiM) direction,

\begin{equation}
\mathbf{t}^{(l)} = 
\boldsymbol{\mu}^{(l)}_{y=1} -
\boldsymbol{\mu}^{(l)}_{y=0},
\end{equation}

where $\boldsymbol{\mu}^{(l)}_y$ is the mean activation of class $y$.

For logistic regression and SVM, the probe direction is given by the learned weight vector. Concept separability is evaluated using AUROC on a train and held-out test split.

\subsection{Activation Steering}
\label{sec:steering}

Following prior work on causal intervention along learned concept directions \cite{arditi2024refusal,shen2020interpreting}, we use activation steering to test whether a biological concept encoded in the Re-ID representation is functionally implicated in identity retrieval. To do so, we introduce a cross-identity retrieval evaluation that measures whether steering a query toward one biological concept systematically shifts its retrieved identities toward that concept.
At layer $l$, we modify the residual-stream representation by adding a scaled concept direction,

\begin{equation}
\mathbf{x}^{(l)\prime} = 
\mathbf{x}^{(l)} \pm
\alpha \mathbf{t}^{(l)},
\label{eq:steering}
\end{equation}

where $\alpha$ controls the intervention strength and the sign of the direction depends on the class we wish to steer towards. The steered activation is added to every token at the layer and the modified activation is propagated through the remaining transformer layers, after which the standard Re-ID embedding and retrieval procedure is applied.

For each identity, we split images into a gallery and query set, excluding same-identity gallery images. We measure how often steering changes a query’s KNN prediction from the same to the opposite concept class, across layers and steering strengths.

Because large interventions can alter the representation without producing a meaningful targeted effect, we additionally introduce an embedding-manifold diagnostic to distinguish concept-directed retrieval changes from representation degradation. We define the manifold range using the 99th percentile of nearest-neighbor distances among gallery embeddings. A steered query whose final embedding falls outside this range is considered degraded.

\subsection{Concept Attribution}
\label{sec:concept_attribution}

We apply representer-point data attribution methods to the probe.
In contrast to most DA work, we refer to the attributed images as constituting the probe's direction. This allows to decompose the probe into a set of smaller, meaningful units (e.g., age–sex class or individuals) for further analysis. Our interpretation follows naturally from the mathematical formulation: for a test activation $\mathbf{x}$, the contribution of training example $i$ is:
\vspace*{-2mm}
\begin{equation}
\tau_i(\mathbf{x}) = 
\lambda_i\mathbf{x}_i^\top\mathbf{x},
\label{eq:representer}
\end{equation}

where $\lambda_i$ is the coefficient associated with training example $i$. This provides a direct decomposition of the probe prediction in terms of training representations.

For SVM probes, the constituting examples are the support vectors. For logistic regression probes, we define the constituting set as the smallest set of examples accounting for 90\% of the total representer-point attribution mass (k90).
We aggregate image-level attributions by individual and quantify enrichment across biological classes to determine which individuals contribute disproportionately to the learned concept.

\subsection{Counterfactual Data Attribution}
\label{sec:loo}

Representer attribution measures how strongly training examples contribute to the learned representation, but contribution to the representation need not equal influence on probe fit. We therefore perform identity-level leave-one-out (LOO) retraining.

For each training individual $j$, we remove all of its images, refit the probe, and measure the resulting change in concept prediction performance:

\begin{equation}
\Delta\mathrm{AUROC}_j = 
\mathrm{AUROC}_{-j}. -
\mathrm{AUROC}_{\mathrm{full}}
\label{eq:loo}
\end{equation}

This quantity measures to what extent concept separability changes when example ($j$) is excluded from training. Together with representer attribution, it reveals to what degree the learned concept direction's predictive performance is sensitive to an individual's removal.

\subsection{Feature-Level Attribution}
\label{sec:feature_attribution}

As demonstrated in DualXDA \cite{yolcu2025sparseefficientexplainabledata}, the representer-point attribution formulation lends itself to relevance propagation via LRP.
This enables us to identify important visual cues underlying highly-attributed images. For the ViT backbone, we use AttnLRP \cite{achtibat2024attnlrp} with CP-LRP \cite{ali2022xai} to obtain pixel-level relevance maps. This provides a localized view of the image regions associated with the learned concept and allows us to inspect potential shortcuts underlying concept predictions.

\section{Experimental Setup}
\label{sec:experimental}

\begin{figure*}[t]
\centering
\begin{subfigure}{0.39\linewidth}
\centering
\includegraphics[width=\linewidth]{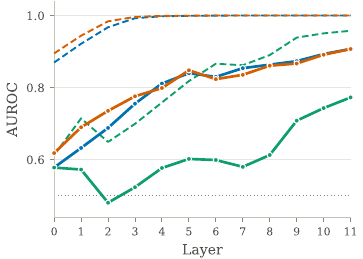}
\caption{Sex}
\label{fig:results_sex_auroc}
\end{subfigure}
\hfill
\begin{subfigure}{0.55\linewidth}
\centering
\includegraphics[width=\linewidth]{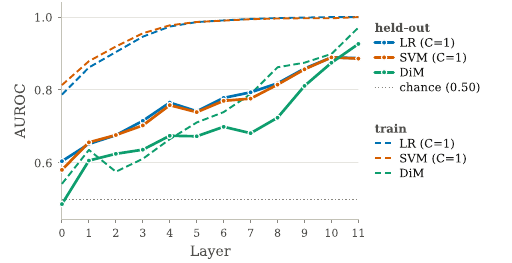}
\caption{Age}
\label{fig:results_age_auroc}
\end{subfigure}
\caption{Layer-wise linear separability of biological concepts in the activations of a ViT fine-tuned for gorilla Re-ID, measured in AUROC.}
\label{fig:both_concepts_auroc}
\end{figure*}

\subsection{Dataset and Concepts}

We use a deduplicated version of \textit{Gorilla-SPAC-Wild} \cite{schall2026gorillawatch}, which contains 45,232 detector-cropped images of 132 individually identified Western lowland gorillas. Identity and age--sex labels \cite{kraus2026age} 
were assigned by expert primatologists, and near-duplicate poses were removed.

From the age--sex labels, we define two binary concepts: \textit{sex} (male/female) and \textit{age} (adult/non-adult). The resulting class composition is given in Table~\ref{tab:dataset_composition}.

\begin{table}[htbp]
  \centering
  \small
  \caption{Dataset composition by age-sex class. Sex is denoted as M/F
  (male/female) and age as A/n-A (adult/non-adult). Silverbacks are sexually
  mature males with silver hair; Blackbacks are maturing males; Adolescents \& Juveniles are non-adults without reliable
  sex traits; and Infants are young dependent individuals.}
  \label{tab:dataset_composition}
  \setlength{\tabcolsep}{3pt}
  \begin{tabular}{lccrrr}
    \toprule
    Class & Sex & Age & Train Ind. & Test Ind. & Images\\
    \midrule
    Silverback                & M & A   & 10 & 6 &  6,477 \\
    Blackback                 & M & A   & 5 & 6 &  2,962 \\
    Adult Female              & F & A   & 31 & 9 & 23,132 \\
    Adolescent \& Juvenile
                              & / & n-A & 21 & 8 & 10,066 \\
    Infant                    & / & n-A & 7 & 3 &  2,595 \\
    \midrule
    Total                     &   &     & 75 & 33 & 45,232 \\
    \bottomrule
  \end{tabular}
\end{table}

To further confirm generalization of our findings on other species, we additionally run our analysis on Jaguar and Polar Bear datasets \cite{rueda2026we, zuerl2023polarbearvidid}.

\subsection{Representations and Concept Probing}

We analyze a DINOv3-Base ViT model before and after fine-tuning for gorilla Re-ID with triplet-margin loss. Images are resized to $224\times224$ pixels. At each transformer layer, we average the $196$ patch-token and $5$ prefix-token activations, obtaining one $768$-dimensional representation per image (we found the choice of which tokens to average over does not matter for probe performance). The fine-tuned model projects its final representation to a $256$-dimensional embedding for identity retrieval.

At every layer, we fit $\ell_2$-regularized logistic regression and linear SVM probes and compare them with the parameter-free difference-in-means direction. Separability is measured by AUROC under two protocols.
Our standard protocol uses the same open-set train/test split as the Re-ID fine-tuning: the test individuals are disjoint from the probe's training individuals and were held out of Re-ID fine-tuning. Probe performance here measures generalization to individuals unseen by both the probe and the Re-ID model.
As a best-case estimate of separability, we additionally run an identity cross-validation (ID-CV) probe that scores each individual using a probe fitted on all images except that individual's.
To test how saliently the concept is represented, we also fit probes using one randomly sampled training image per identity. We repeat this sampling 32 times and report the mean and standard deviation across repetitions.

\subsection{Activation Steering}

We test the functional relevance of each unit-normalized probe direction by intervening at every transformer layer over a range of steering strengths $\alpha$, as defined in Eq.~\ref{eq:steering}. After the intervention, we propagate the modified activations through the remaining layers and perform cosine-similarity retrieval in the $256$-dimensional Re-ID embedding space.

For each identity, we build a fixed split with 80\% of its images forming the (unsteered) gallery and 20\% forming the query set. When steering a query, we exclude gallery images of the same individual and predict the identity that is the majority among the $K$ nearest gallery images.
Excluding same-ID gallery images is important, as they cluster so tightly in the embedding space that steering would otherwise become impossible. KNN-predictions on this cross-identity gallery thus ask which \textit{other} gorilla is the most similar to the query. For our main experiments, we sweep across $\alpha \in \{0, 1, 2, 4, 8, 16, 32, 64\}$ (0 is the baseline no-steering effect) and use $K=5$, though we show that other values exhibit the same effect.

Our main metric to measure steering effect considers the percentage of queries whose KNN prediction \textit{flips}, from having the same concept label (e.g., same sex) when not steering to the opposite label, upon steering.
To distinguish targeted steering effects from mere representation degradation, we introduce the following diagnostic: we calculate the 99th percentile of the distances between each gallery embedding and its closest neighbor and define this distance as the range in which regular embeddings reside. A steered query leading to a final-layer embedding outside this manifold (i.e., having its nearest gallery neighbor further away than that 99th percentile distance) is deemed degraded and thus not meaningful.

\subsection{Data and Feature Attribution}

To gain deeper insight into the representation we found, we decompose the probe
into its constituent training images using representer-point data attribution and cross-check identity-level influence via ID-leave-one-out retraining. For SVM probes, the constituting set comprises the support vectors. For logistic-regression probes, it is the smallest set of training images accounting for 90\% of the total representer-attribution mass (k90). We aggregate image-level attributions by identity and report the enrichment of each age--sex class and identity relative to its training-set prevalence.

Finally, for selected high-attribution test--training pairs, we use DualXDA \cite{yolcu2025sparseefficientexplainabledata} to propagate attribution to both input images. We obtain pixel-level relevance maps with AttnLRP \cite{achtibat2024attnlrp} and CP-LRP \cite{ali2022xai}, using the \texttt{lxt} implementation.

\section{Results}
\label{sec:results}

\begin{figure}[t]
\centering
\includegraphics[width=0.8\linewidth]{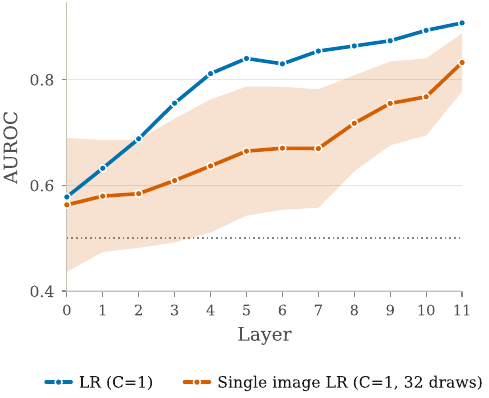}
\caption{Concept separability with the full training set versus one image per individual. Shading indicates one standard deviation across 32 random single-image samples.}
\label{fig:single_image_probe}
\end{figure}

\subsection{Re-ID features encode biological concepts}
\label{sec:results_linear}

We first ask whether a Re-ID model trained solely to distinguish individuals develops linearly decodable biological concepts. We find that it does. A linear probe trained on the residual-stream activations of the Re-ID-fine-tuned DINOv3 model attains a peak image-level AUROC of 0.91 for sex and 0.89 for age on held-out individuals. Aggregating predictions to measure identity-level AUROC results in values of up to 0.97. The corresponding training-set AUROCs reach 1.00, indicating that the training set is perfectly separable.

The representation becomes increasingly linearly separable in the later layers (Fig.~\ref{fig:both_concepts_auroc}). Logistic regression and SVM probes produce nearly identical results, while the parameter-free difference-in-means direction performs substantially worse (0.77 AUROC for sex), indicating that the concepts are encoded linearly but are not simply captured by the difference between class means. We therefore use the logistic regression for the subsequent causal experiments; corresponding SVM and age results are provided in the appendix.

Importantly, the concept remains recoverable when the probe receives only one image per individual (Fig.~\ref{fig:single_image_probe}). Despite being trained on only 46 images, the single-image-per-ID probes recover strong separation in the middle and late layers across 32 random samplings. This demonstrates that the concept is not dependent on dense observations of particular individuals, but is sufficiently structured to be recovered from sparse individual-level supervision.

We additionally evaluated jaguar and polar bear Re-ID datasets and observed similar concept recoverability, while also finding that performance depends on the visual distinguishability of the corresponding biological traits.  

\subsection{Re-ID fine-tuning reshapes biological concepts}
\label{sec:results_finetuning}

\begin{figure}[t]
\centering
\includegraphics[width=0.8\linewidth]{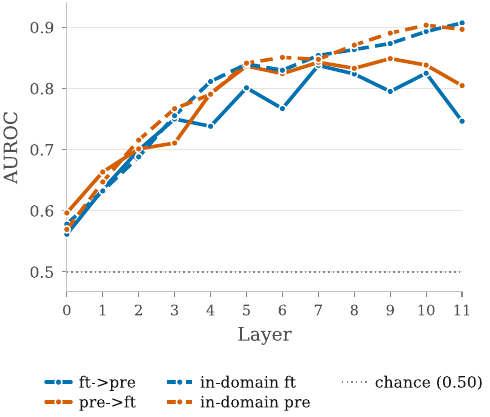}
\caption{Cross-model transfer between probes trained on fine-tuned activations (ft$\rightarrow$pre) and pre-trained-only activations (pre$\rightarrow$ft). In-domain probes show the normal probes' AUROCs.}
\label{fig:finetuning_results}
\end{figure}

\begin{figure*}[t]
\centering
\begin{subfigure}{0.4\linewidth}
\centering
\includegraphics[width=\linewidth]{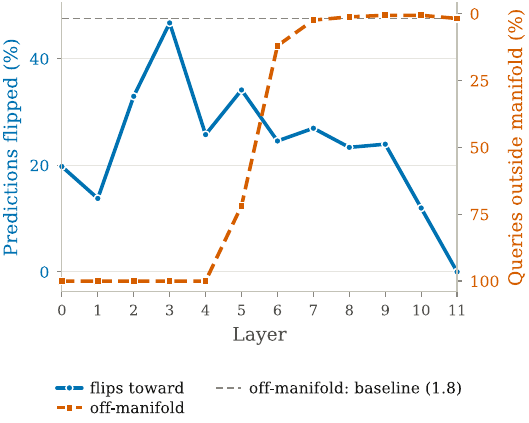}
\caption{Steering effect vs degradation at $\alpha=16$ by layer.}
\label{fig:steering_by_layer}
\end{subfigure}
\hfill
\begin{subfigure}{0.4\linewidth}
\centering
\includegraphics[width=\linewidth]{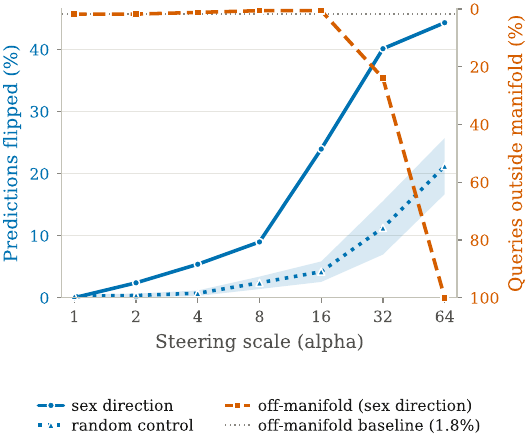}
\caption{Steering effect by $\alpha$ at layer 9, with random control.}
\label{fig:steering_by_alpha}
\end{subfigure}
\caption{Percentage of steered queries flipping their prediction towards an individual of the opposite sex compared to the baseline. Orange shows the embedding degradation as percentage of steered queries leaving the defined embedding manifold. The layer-wise plot shows that steering only becomes meaningful at middle layers. Increasing steering dose ($\alpha$) produces more steering effect but also higher degradation.}
\label{fig:steering_dose}
\end{figure*}

We next ask whether the biological concepts are created by Re-ID fine-tuning or inherited from the pretrained visual representation. Sex is already strongly linearly separable in the stock DINOv3 backbone, with AUROCs comparable to those of the Re-ID-fine-tuned model. Thus, Re-ID training does not create the sex representation from scratch.

The geometry of the representation nevertheless changes during fine-tuning. Probe directions learned independently on pretrained and fine-tuned activations become increasingly non-collinear in later layers, with cosine similarity gradually decreasing from around 0.95 in the first layer to 0.25 in the final layer. However, cosine similarity alone does not establish functional divergence because fine-tuning may induce activation space transformations that preserve the underlying concept. We therefore evaluate cross-model probe transfer. Probes trained on one representation are applied directly to the other and compared with in-domain probes (Fig.~\ref{fig:finetuning_results}). 
Transfer remains strong through most layers but degrades near the model output. The overall pattern is consistent across sex and age, although age exhibits asymmetric transfer in the middle layers.
This suggests that Re-ID fine-tuning largely preserves the biological concept geometry through most of the network while reshaping its representation near the task-specific output. Thus, fine-tuning does not simply introduce biological concepts; rather, it modifies an existing semantic representation to support identity discrimination.

\subsection{Steering reveals causal use of the sex direction}
\label{sec:results_steering}

Linear decodability shows that sex information is present in the model's representation, but it does not tell us whether the model actually uses this information for Re-ID. To test this, we intervene on the learned sex direction and examine how the model's identity predictions change (Fig.~\ref{fig:steering_dose}). 

Steering each query's representation toward the opposite sex shifts the model's cross-identity KNN predictions toward individuals of that sex. At an intervention strength of $\alpha=16$, we find that 24\% of queries flipped their predicted ID from the same sex (at baseline $\alpha=0$) toward the opposite, with less than 1\% of queries leaving our defined embedding region.
To confirm that the effect stems from the direction's content rather than just the perturbation itself, we steer along random directions (5 seeds), which yields a flip-toward value of $<$4\%.
Stronger interventions produce larger apparent effects but also cause the representation to collapse, with nearly all queries leaving the manifold.

Our results demonstrate a causal link between the learned sex representation and Re-ID predictions: 
as the intervention acts at an intermediate layer and the flip is read from the final retrieval, the effect propagates through the forward pass to change the model's actual prediction.
Importantly, the effect appears well before severe representation degradation sets in. It reflects the model's use of sex information rather than an artifact of extreme perturbation.

\subsection{The learned sex representation forms a graded biological axis}
\label{sec:results_biology}

\begin{figure}[t]
\centering
\includegraphics[width=0.9\linewidth]{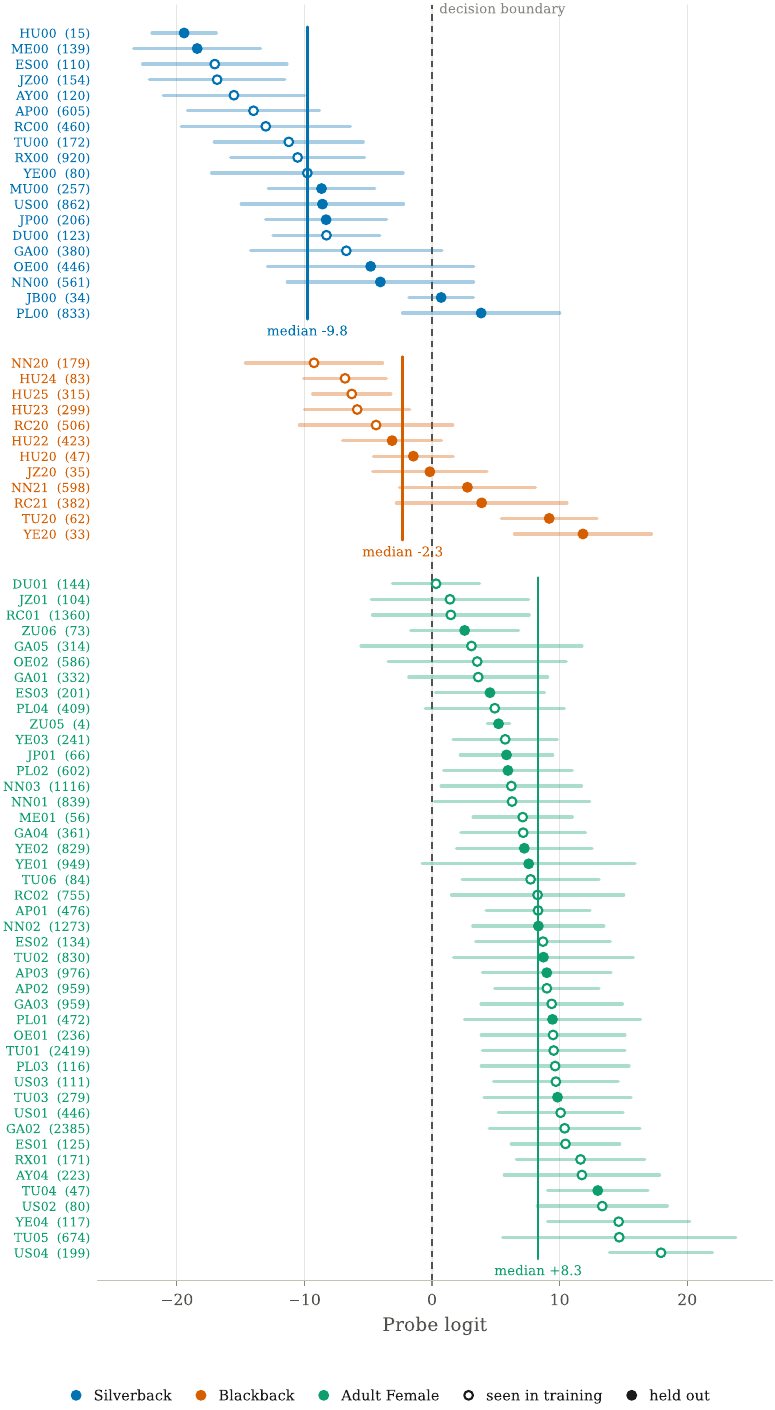}
\caption{The ID-CV probe's mean logit for each held-out individual. Individuals form a graded axis, with silverbacks and adult females occupying opposite extremes and blackbacks concentrated between them.}
\label{fig:logit_per_identity}
\end{figure}

Having established that sex is both linearly encoded and causally used, we next examine what structure the learned direction captures. Rather than forming a binary male--female separation, the ID-cross-validation probe scores form a graded axis (Fig.~\ref{fig:logit_per_identity}). Most silverbacks and adult females occupy opposite ends of the axis, while blackbacks are concentrated between them.

This structure is consistent with the biological development of sexual dimorphism in gorillas: visual sex-status classification becomes reliable only with maturity, and blackbacks can remain difficult to distinguish from adult females \cite{robbins2022population}. The representation therefore appears to reflect a biologically meaningful continuum rather than merely reproducing the binary labels used to train the probe.

\subsection{Difficult individuals shape the direction, but no individual is essential}
\label{sec:results_attribution}

\begin{figure}[t]
\centering
\includegraphics[width=0.75\linewidth]{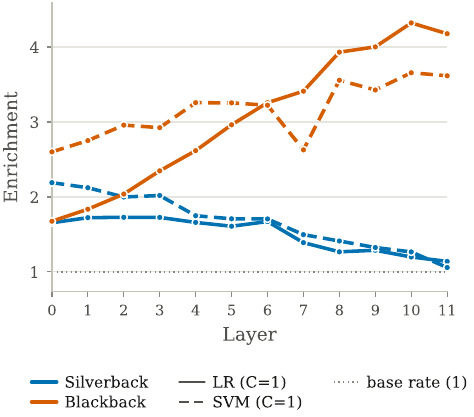}
\caption{Enrichment of silverbacks and blackbacks among the constituting examples of the sex probe. Enrichment is defined as the ratio (share in the SVs/k90 / share in the training pool). Blackbacks become increasingly over-represented in deeper layers.}
\label{fig:male_share_top}
\end{figure}

The ID-CV probe's logit-level analysis shows that blackback individuals cluster near the decision boundary. We further apply representer-point DA to identify training images driving the probe. Blackbacks are enriched 4$\times$ among the highest-attributing images (support vectors / k90) relative to their dataset share (Fig.~\ref{fig:male_share_top}). The same pattern holds at the individual level: gorillas contributing most to the highest-attributing samples are disproportionately blackbacks.

This enrichment follows naturally from the graded representation: prototypical silverbacks and adult females lie far from the decision boundary and therefore contribute relatively little to defining it, whereas visually ambiguous blackbacks occupy the boundary region. The probe is thus disproportionately constituted by those difficult blackbacks.

However, disproportionally constituting the direction does not imply that any single individual is necessary for the representation. Exact identity-level LOO retraining shows that removing single individuals changes held-out AUROC by up to 0.016.
Thus, despite the strong enrichment of difficult individuals among the constituting examples, the biological concept is redundantly encoded across the population.
We also find that individuals ranked highest by the representer-point methods are also ranked highest in LOO-influence, though some of them have positive $\Delta$AUROC, i.e., their removal would \emph{improve} the probe.
This illustrates that constitution and LOO-influence are distinct. 

Lastly, propagating the SVM probe's influences back to produce pixel-level relevance maps via DualXDA, we find relevance concentrates on the face and sex-associated anatomical regions for most images. Among images showing adult female individuals nursing an infant, relevance often localizes to the infant as well, highlighting a potential shortcut in sex prediction.
\section{Conclusion}
\label{sec:conclusion}

We investigated whether a ViT trained for animal Re-ID develops interpretable biological concepts despite being optimized solely for individual identity matching. We find that sex and age concepts emerge as linearly recoverable directions that generalize to unseen individuals and remain detectable from a single image per individual, indicating structured rather than memorized information. The sex direction is also functionally implicated in retrieval: intermediate-strength steering shifts cross-identity predictions toward the opposite sex, while stronger interventions cause representation collapse. Re-ID fine-tuning does not create these concepts but largely preserves their geometry while reshaping it near the output.

Data attribution further reveals a graded sex axis, with visually ambiguous individuals disproportionately constituting the direction, yet no individual being essential to its performance. Importantly, constitution and counterfactual influence do not necessarily coincide. Feature attribution further exposes potential shortcuts. For example, relevance for females sometimes localizing on the infant. 

Together, these findings demonstrate that interpretability can expose both the semantic structure and potential failure modes of animal Re-ID models. In a conservation setting, such tools help to determine not only \emph{whether} a model recognizes an individual, but also \emph{what biological structure it has learned, how that structure is used, which training examples shape it, and where it may rely on spurious cues}. This provides a step toward more auditable and scientifically trustworthy systems for wildlife monitoring.

{
    \small
    \bibliographystyle{ieeenat_fullname}
    \bibliography{main}

\begin{thebibliography}{32}
\providecommand{\natexlab}[1]{#1}
\providecommand{\url}[1]{\texttt{#1}}
\expandafter\ifx\csname urlstyle\endcsname\relax
  \providecommand{\doi}[1]{doi: #1}\else
  \providecommand{\doi}{doi: \begingroup \urlstyle{rm}\Url}\fi

\bibitem[Achtibat et~al.(2024)Achtibat, Hatefi, Dreyer, Jain, Wiegand,
  Lapuschkin, and Samek]{achtibat2024attnlrp}
Reduan Achtibat, Sayed Mohammad~Vakilzadeh Hatefi, Maximilian Dreyer, Aakriti
  Jain, Thomas Wiegand, Sebastian Lapuschkin, and Wojciech Samek.
\newblock Attnlrp: Attention-aware layer-wise relevance propagation for
  transformers.
\newblock \emph{arXiv preprint arXiv:2402.05602}, 2024.

\bibitem[Alain and Bengio(2016)]{alain2016understanding}
Guillaume Alain and Yoshua Bengio.
\newblock Understanding intermediate layers using linear classifier probes.
\newblock \emph{arXiv preprint arXiv:1610.01644}, 2016.

\bibitem[Ali et~al.(2022)Ali, Schnake, Eberle, Montavon, M{\"u}ller, and
  Wolf]{ali2022xai}
Ameen Ali, Thomas Schnake, Oliver Eberle, Gr{\'e}goire Montavon, Klaus-Robert
  M{\"u}ller, and Lior Wolf.
\newblock Xai for transformers: Better explanations through conservative
  propagation.
\newblock In \emph{International conference on machine learning}, pages
  435--451. PMLR, 2022.

\bibitem[Arditi et~al.(2024)Arditi, Obeso, Syed, Paleka, Panickssery, Gurnee,
  and Nanda]{arditi2024refusal}
Andy Arditi, Oscar Obeso, Aaquib Syed, Daniel Paleka, Nina Panickssery, Wes
  Gurnee, and Neel Nanda.
\newblock Refusal in language models is mediated by a single direction.
\newblock \emph{Advances in Neural Information Processing Systems},
  37:\penalty0 136037--136083, 2024.

\bibitem[Bergamini et~al.(2018)Bergamini, Porrello, Dondona, Del~Negro,
  Mattioli, D'alterio, and Calderara]{bergamini2018multi}
Luca Bergamini, Angelo Porrello, Andrea~Capobianco Dondona, Ercole Del~Negro,
  Mauro Mattioli, Nicola D'alterio, and Simone Calderara.
\newblock Multi-views embedding for cattle re-identification.
\newblock In \emph{2018 14th international conference on signal-image
  technology \& internet-based systems (SITIS)}, pages 184--191. IEEE, 2018.

\bibitem[Chen et~al.(2020)Chen, Swarup, Matkowski, Kong, Han, Zhang, and
  Rong]{chen2020panda}
Peng Chen, Pranjal Swarup, Wojciech~Michal Matkowski, Adams Wai~Kin Kong, Su
  Han, Zhihe Zhang, and Hou Rong.
\newblock A study on giant panda recognition based on images of a large
  proportion of captive pandas.
\newblock \emph{Ecology and Evolution}, 10\penalty0 (7):\penalty0 3561--3573,
  2020.

\bibitem[Clapham et~al.(2020)Clapham, Miller, Nguyen, and
  Darimont]{clapham2020bear}
Melanie Clapham, Ed Miller, Mary Nguyen, and Chris~T. Darimont.
\newblock Automated facial recognition for wildlife that lack unique markings:
  A deep learning approach for brown bears.
\newblock \emph{Ecology and Evolution}, 10\penalty0 (23):\penalty0
  12883--12892, 2020.

\bibitem[Dosovitskiy et~al.(2020)Dosovitskiy, Beyer, Kolesnikov, Weissenborn,
  Zhai, Unterthiner, Dehghani, Minderer, Heigold, Gelly,
  et~al.]{dosovitskiy2020image}
Alexey Dosovitskiy, Lucas Beyer, Alexander Kolesnikov, Dirk Weissenborn,
  Xiaohua Zhai, Thomas Unterthiner, Mostafa Dehghani, Matthias Minderer, Georg
  Heigold, Sylvain Gelly, et~al.
\newblock An image is worth 16x16 words: Transformers for image recognition at
  scale.
\newblock \emph{arXiv preprint arXiv:2010.11929}, 2020.

\bibitem[Gurnee and Tegmark(2024)]{gurnee2024language}
Wes Gurnee and Max Tegmark.
\newblock Language models represent space and time.
\newblock In \emph{International Conference on Learning Representations}, pages
  2483--2503, 2024.

\bibitem[He et~al.(2021)He, Luo, Wang, Wang, Li, and Jiang]{he2021transreid}
Shuting He, Hao Luo, Pichao Wang, Fan Wang, Hao Li, and Wei Jiang.
\newblock Transreid: Transformer-based object re-identification.
\newblock In \emph{Proceedings of the IEEE/CVF international conference on
  computer vision}, pages 15013--15022, 2021.

\bibitem[Hermans et~al.(2017)Hermans, Beyer, and Leibe]{hermans2017defense}
Alexander Hermans, Lucas Beyer, and Bastian Leibe.
\newblock In defense of the triplet loss for person re-identification.
\newblock \emph{arXiv preprint arXiv:1703.07737}, 2017.

\bibitem[Hinton(1986)]{hinton1986learning}
Geoffrey~E Hinton.
\newblock Learning distributed representations of concepts.
\newblock In \emph{Proceedings of the Annual Meeting of the Cognitive Science
  Society}, 1986.

\bibitem[Kim et~al.(2018)Kim, Wattenberg, Gilmer, Cai, Wexler, Viegas,
  et~al.]{kim2018interpretability}
Been Kim, Martin Wattenberg, Justin Gilmer, Carrie Cai, James Wexler, Fernanda
  Viegas, et~al.
\newblock Interpretability beyond feature attribution: Quantitative testing
  with concept activation vectors (tcav).
\newblock In \emph{International conference on machine learning}, pages
  2668--2677. PMLR, 2018.

\bibitem[Koh and Liang(2017)]{koh2017understanding}
Pang~Wei Koh and Percy Liang.
\newblock Understanding black-box predictions via influence functions.
\newblock In \emph{International conference on machine learning}, pages
  1885--1894. PMLR, 2017.

\bibitem[Konz et~al.(2023)Konz, Godfrey, Shapiro, Tu, Kvinge, and
  Brown]{konz2023attributing}
Nicholas Konz, Charles Godfrey, Madelyn Shapiro, Jonathan Tu, Henry Kvinge, and
  Davis Brown.
\newblock Attributing learned concepts in neural networks to training data.
\newblock \emph{arXiv preprint arXiv:2310.03149}, 2023.

\bibitem[Kraus et~al.(2026)Kraus, Huang, Li, Cui, Wang, Li, Liu, Wang, Strier,
  and Xiao]{kraus2026age}
Jacob~B Kraus, Zhi~Pang Huang, Yan~Pang Li, Liang~Wei Cui, Shuang~Jin Wang,
  Jin~Fa Li, Feng Liu, Yun Wang, Karen~B Strier, and Wen Xiao.
\newblock Age--sex class variation in the activity budget and diet of
  rhinopithecus bieti in association with monthly temperature: Jb kraus et al.
\newblock \emph{International Journal of Primatology}, pages 1--27, 2026.

\bibitem[Laskowski et~al.(2023)Laskowski, Sawahn, Schall, Wasmuht, Bermejo, and
  de~Melo]{laskowski2023gorillavision}
Lukas Laskowski, Rohan Sawahn, Maximilian Schall, Dante Wasmuht, Magdalena
  Bermejo, and Gerard de Melo.
\newblock Gorillavision--open-set reidentification of wild gorillas.
\newblock \emph{CamTrap WS}, 23, 2023.

\bibitem[Marks and Tegmark(2023)]{marks2023geometry}
Samuel Marks and Max Tegmark.
\newblock The geometry of truth: Emergent linear structure in large language
  model representations of true/false datasets.
\newblock \emph{arXiv preprint arXiv:2310.06824}, 2023.

\bibitem[Mikolov et~al.(2013)Mikolov, Chen, Corrado, and
  Dean]{mikolov2013efficientestimationwordrepresentations}
Tomas Mikolov, Kai Chen, Greg Corrado, and Jeffrey Dean.
\newblock Efficient estimation of word representations in vector space, 2013.

\bibitem[Norouzzadeh et~al.(2018)Norouzzadeh, Nguyen, Kosmala, Swanson, Palmer,
  Packer, and Clune]{norouzzadeh2018automatically}
Mohammad~Sadegh Norouzzadeh, Anh Nguyen, Margaret Kosmala, Alexandra Swanson,
  Meredith~S Palmer, Craig Packer, and Jeff Clune.
\newblock Automatically identifying, counting, and describing wild animals in
  camera-trap images with deep learning.
\newblock \emph{Proceedings of the National Academy of Sciences}, 115\penalty0
  (25):\penalty0 E5716--E5725, 2018.

\bibitem[Oquab et~al.(2023)Oquab, Darcet, Moutakanni, Vo, Szafraniec, Khalidov,
  Fernandez, Haziza, Massa, El-Nouby, et~al.]{oquab2023dinov2}
Maxime Oquab, Timoth{\'e}e Darcet, Th{\'e}o Moutakanni, Huy Vo, Marc
  Szafraniec, Vasil Khalidov, Pierre Fernandez, Daniel Haziza, Francisco Massa,
  Alaaeldin El-Nouby, et~al.
\newblock Dinov2: Learning robust visual features without supervision.
\newblock \emph{arXiv preprint arXiv:2304.07193}, 2023.

\bibitem[Park et~al.(2023{\natexlab{a}})Park, Choe, and Veitch]{park2023linear}
Kiho Park, Yo~Joong Choe, and Victor Veitch.
\newblock The linear representation hypothesis and the geometry of large
  language models.
\newblock \emph{arXiv preprint arXiv:2311.03658}, 2023{\natexlab{a}}.

\bibitem[Park et~al.(2023{\natexlab{b}})Park, Georgiev, Ilyas, Leclerc, and
  Madry]{park2023trak}
Sung~Min Park, Kristian Georgiev, Andrew Ilyas, Guillaume Leclerc, and
  Aleksander Madry.
\newblock Trak: Attributing model behavior at scale.
\newblock \emph{arXiv preprint arXiv:2303.14186}, 2023{\natexlab{b}}.

\bibitem[Ravichander et~al.(2021)Ravichander, Belinkov, and
  Hovy]{ravichander2021probing}
Abhilasha Ravichander, Yonatan Belinkov, and Eduard Hovy.
\newblock Probing the probing paradigm: Does probing accuracy entail task
  relevance?
\newblock In \emph{Proceedings of the 16th Conference of the European Chapter
  of the Association for Computational Linguistics: Main Volume}, pages
  3363--3377, 2021.

\bibitem[Robbins et~al.(2022)Robbins, Manguette, Breuer, Groenenberg, Parnell,
  Stephan, Stokes, and Robbins]{robbins2022population}
Andrew~M Robbins, Marie~L Manguette, Thomas Breuer, Milou Groenenberg,
  Richard~J Parnell, Claudia Stephan, Emma~J Stokes, and Martha~M Robbins.
\newblock Population dynamics of western gorillas at mbeli bai.
\newblock \emph{Plos one}, 17\penalty0 (10):\penalty0 e0275635, 2022.

\bibitem[Rueda-Toicen et~al.(2026)Rueda-Toicen, Martin, Morozov, Mahmood,
  Schild, Dayani, Panza, and de~Melo]{rueda2026we}
Antonio Rueda-Toicen, Abigail~Allen Martin, Daniil Morozov, Matin Mahmood,
  Alexandra Schild, Shahabeddin Dayani, Davide Panza, and Gerard de Melo.
\newblock Are we recognizing the jaguar or its background? a diagnostic
  framework for jaguar re-identification.
\newblock \emph{arXiv preprint arXiv:2604.09690}, 2026.

\bibitem[Schall et~al.(2026)Schall, Kn{\"o}fel, K{\"o}nig, Kubeler, von
  Klinski, Linnemann, Liu, Schlegelmilch, Woyciniuk, Schild,
  et~al.]{schall2026gorillawatch}
Maximilian Schall, Felix~Leonard Kn{\"o}fel, Noah~Elias K{\"o}nig, Jan~Jonas
  Kubeler, Maximilian von Klinski, Joan~Wilhelm Linnemann, Xiaoshi Liu,
  Iven~Jelle Schlegelmilch, Ole Woyciniuk, Alexandra Schild, et~al.
\newblock Gorillawatch: An automated system for in-the-wild gorilla
  re-identification and population monitoring.
\newblock In \emph{Proceedings of the IEEE/CVF Winter Conference on
  Applications of Computer Vision}, pages 8364--8375, 2026.

\bibitem[Shen et~al.(2020)Shen, Gu, Tang, and Zhou]{shen2020interpreting}
Yujun Shen, Jinjin Gu, Xiaoou Tang, and Bolei Zhou.
\newblock Interpreting the latent space of gans for semantic face editing.
\newblock In \emph{Proceedings of the IEEE/CVF conference on computer vision
  and pattern recognition}, pages 9243--9252, 2020.

\bibitem[Siméoni et~al.(2025)Siméoni, Vo, Seitzer, Baldassarre, Oquab, Jose,
  Khalidov, Szafraniec, Yi, Ramamonjisoa, Massa, Haziza, Wehrstedt, Wang,
  Darcet, Moutakanni, Sentana, Roberts, Vedaldi, Tolan, Brandt, Couprie,
  Mairal, Jégou, Labatut, and Bojanowski]{simeoni2025dinov3}
Oriane Siméoni, Huy~V. Vo, Maximilian Seitzer, Federico Baldassarre, Maxime
  Oquab, Cijo Jose, Vasil Khalidov, Marc Szafraniec, Seungeun Yi, Michaël
  Ramamonjisoa, Francisco Massa, Daniel Haziza, Luca Wehrstedt, Jianyuan Wang,
  Timothée Darcet, Théo Moutakanni, Leonel Sentana, Claire Roberts, Andrea
  Vedaldi, Jamie Tolan, John Brandt, Camille Couprie, Julien Mairal, Hervé
  Jégou, Patrick Labatut, and Piotr Bojanowski.
\newblock Dinov3, 2025.

\bibitem[Yeh et~al.(2018)Yeh, Kim, Yen, and Ravikumar]{yeh2018representer}
Chih-Kuan Yeh, Joon Kim, Ian En-Hsu Yen, and Pradeep~K Ravikumar.
\newblock Representer point selection for explaining deep neural networks.
\newblock \emph{Advances in neural information processing systems}, 31, 2018.

\bibitem[Zuerl et~al.(2023)Zuerl, Dirauf, Koeferl, Steinlein, Sueskind, Zanca,
  Brehm, Fersen, and Eskofier]{zuerl2023polarbearvidid}
Matthias Zuerl, Richard Dirauf, Franz Koeferl, Nils Steinlein, Jonas Sueskind,
  Dario Zanca, Ingrid Brehm, Lorenzo~von Fersen, and Bjoern Eskofier.
\newblock Polarbearvidid: A video-based re-identification benchmark dataset for
  polar bears.
\newblock \emph{Animals}, 13\penalty0 (5):\penalty0 801, 2023.

\bibitem[Ümit Yolcu et~al.(2025)Ümit Yolcu, Weckbecker, Wiegand, Samek, and
  Lapuschkin]{yolcu2025sparseefficientexplainabledata}
Galip Ümit Yolcu, Moritz Weckbecker, Thomas Wiegand, Wojciech Samek, and
  Sebastian Lapuschkin.
\newblock Sparse, efficient and explainable data attribution with dualxda,
  2025.

\end{thebibliography}
}

\clearpage
\appendix
\maketitlesupplementary

\section{Limitations and Future Work}

Our study focuses primarily on age and sex in gorillas because concept-level analyses of animal Re-ID models require an uncommon combination of annotations: a sufficiently large number of known identities, multiple images per identity, and reliable concept labels. Few currently available wildlife datasets satisfy all three requirements. The gorilla sex--age dataset therefore provides a particularly suitable testbed for controlled layer-wise, subspace, and causal analyses. We complement these experiments with checks on jaguars \cite{rueda2026we} and polar bears \cite{zuerl2023polarbearvidid}, providing initial evidence that the observed phenomena are not restricted to a single species. Broader validation across species, datasets, and Re-ID architectures will be important as suitably annotated data become available. Our framework provides a direct basis for such comparative studies.

Our analysis considers two high-level biological concepts, whereas animal Re-ID relies on a broader range of biological and visual cues, including coat color, texture, markings, body shape, and pose. Extending the analysis across abstraction levels could reveal a richer hierarchy of representations, including lower-level concepts that emerge in earlier layers. This requires datasets combining identity labels with richer concept annotations to disentangle target concepts such as age and sex from correlated cues such as body size, morphology, and appearance.

\section{Additional Results} 

\subsection{Age Concept} \label{app:age}

\begin{figure} [t]
\centering
\begin{subfigure}{0.87\linewidth}
    \centering
    \includegraphics[width=\linewidth]{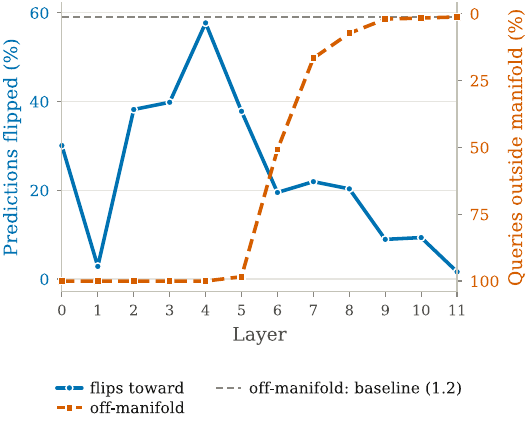}
    \caption{Steering effect vs degradation at $\alpha=16$ by layer.}
    \label{fig:age_steering_by_layer}
\end{subfigure}
\medskip
\begin{subfigure}{0.87\linewidth}
    \centering
    \includegraphics[width=\linewidth]{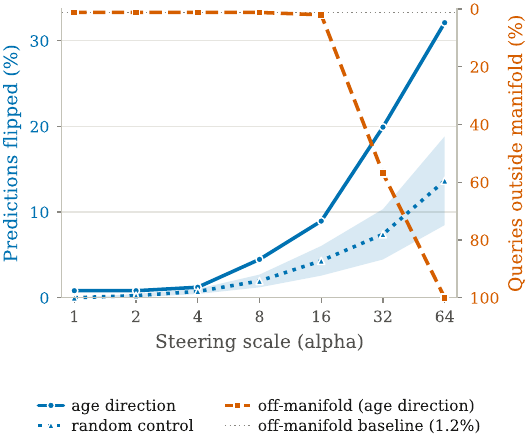}
    \caption{Predicted age label flips by $\alpha$ at layer 9 with random control.} 
    \label{fig:age_steering_by_alpha}
\end{subfigure}
\caption{Percentage of steered queries that switch to an individual of the opposite age. Orange denotes off-manifold degradation. Steering becomes effective in the middle layers and increases with $\alpha$, alongside greater degradation.}
\label{fig:age_steering}
\end{figure}

In the main paper, we mostly focused on sex concept, as age exhibits broadly similar qualitative behavior. In the following section, we present selected results for age, emphasizing the differences observed between the two concepts.

Despite comparable probe AUROCs, steering towards the opposite age was less effective than for the sex concept (Figs.~\ref{fig:age_steering_by_layer} and~\ref{fig:age_steering_by_alpha}). Substantial steering effects emerged only once off-manifold degradation had already become pronounced. Further analysis showed that individuals sharing the same age label formed tighter clusters in the embedding space. Consequently, steering representations out of these clusters frequently resulted in representation collapse.

\begin{figure}[t]
\centering
\includegraphics[width=0.83\linewidth]{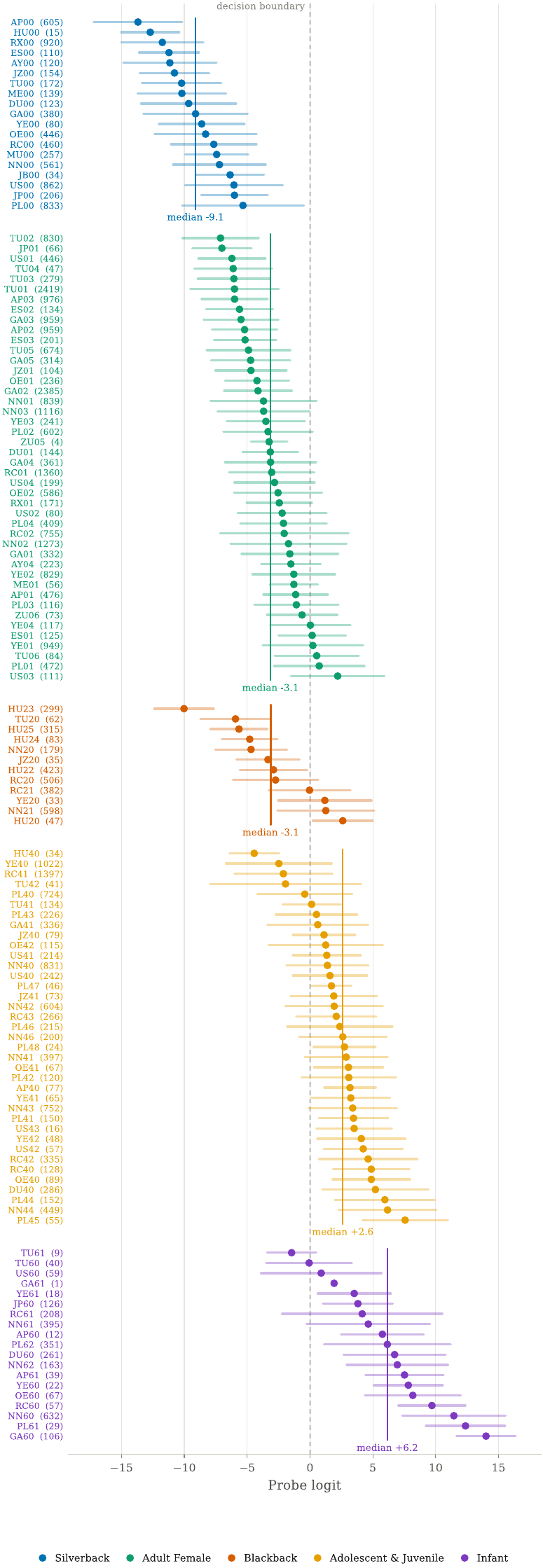}
\caption{ID-CV probe's mean age-probe logit for each held-out individual. The same graded axis as with the sex-probe emerges, with the distinctive silverbacks and infants making up the extremes.}
\label{fig:age_logit_per_identity}
\end{figure}

ID-CV probes trained for the age concept exhibited a similar gradation in their logits. Age-ambiguous blackbacks and individuals labeled as Adolescent \& Juvenile were concentrated near the decision boundary (Fig.~\ref{fig:age_logit_per_identity}).

\begin{figure*}[t]
\centering
\includegraphics[width=\textwidth]{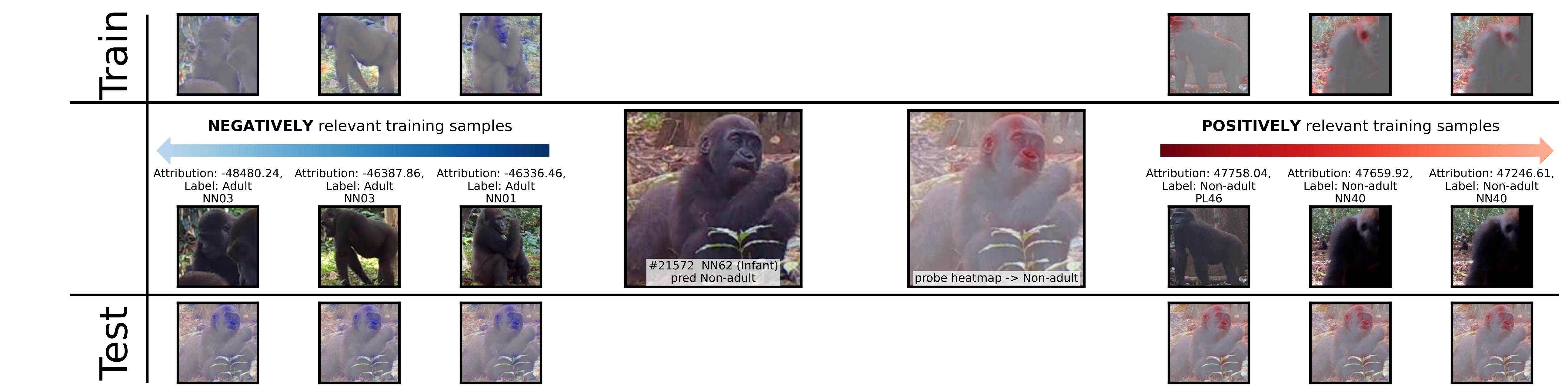}
\caption{DualXDA heatmap for the 'age' probe, evaluated on an infant gorilla. The same Re-ID relevant regions are highlighted.}
\label{fig:dualxda_infant}
\end{figure*}

Finally, the DualXDA heatmaps showed spatial relevance patterns similar to those observed for the sex concept, primarily highlighting common Re-ID cues such as the face and other anatomical regions (Fig.~\ref{fig:dualxda_infant}).

\subsection{Concepts in Representations of Other Species}
\label{app:other_species}

\begin{figure}[t]
\centering
\begin{subfigure}{0.9\linewidth}
    \centering
    \includegraphics[width=\linewidth]{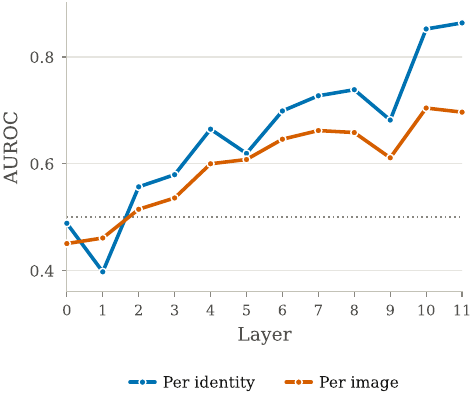}
    \caption{Jaguars}
    \label{fig:jaguar_auroc}
\end{subfigure}
\medskip
\begin{subfigure}{0.9\linewidth}
    \centering
    \includegraphics[width=\linewidth]{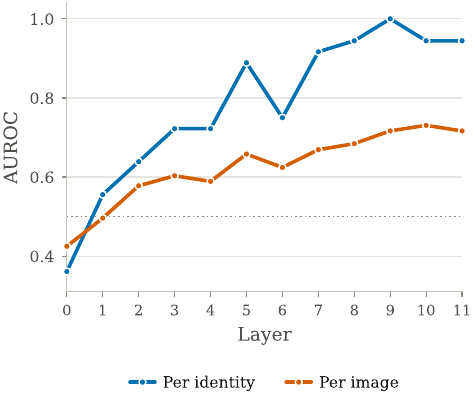}
    \caption{Polar bears}
    \label{fig:polar_auroc}
\end{subfigure}
\caption{Layer-wise AUROCs of LR sex probes on the jaguar and polar bear datasets, with one identity or one zoo held out. ``Per identity'' denotes aggregating logits over all images of an identity and evaluating the resulting identity-level predictions.}
\label{fig:other_datasets_auroc}
\end{figure}

We further examined whether the observed layer-wise behavior generalizes to Re-ID datasets of other species. However, datasets combining a sufficient number of identities with sex or age annotations are scarce. We identified two suitable datasets, summarized in Table~\ref{tab:other_datasets}. Given their limited number of identities, we train ID-CV probes rather than using fixed held-out splits (Fig.~\ref{fig:other_datasets_auroc}).

\begin{table}
\centering
\caption{Statistics of the additional Re-ID datasets. A dash indicates unavailable information.}
\label{tab:other_datasets}
\small
\setlength{\tabcolsep}{3.5pt}
\begin{tabular}{lccccc}
\toprule
Dataset & Images & IDs & Male & Female \\
\midrule
Jaguar \cite{rueda2026we} & 1.861 & 30 & 8 & 22 \\
PolarBearVidID \cite{zuerl2023polarbearvidid} & 138.363 & 13 & 4 & 9 \\
\bottomrule
\end{tabular}
\end{table}

On the jaguar dataset, the sex probe reaches a peak AUROC of 0.864 and reproduces the trend observed in gorillas, with sex becoming increasingly separable in the later layers.

PolarBearVidID presents an additional challenge because enclosure background is confounded with sex: two zoos contain only female individuals. Under leave-one-identity-out evaluation, the probe may therefore exploit zoo-specific background cues shared between the training and test sets. To mitigate this confound, we additionally perform leave-one-zoo-out evaluation. This setting yields a peak AUROC of 1.0; however, the corresponding held-out set contains only two individuals. This result should therefore be interpreted as preliminary evidence rather than a robust estimate of cross-zoo generalization.

\section{Additional Analyses}
\label{app:additional_analyses}

In this section, we evaluate the robustness of our method. Following the main paper, all analyses use layer 9. We select this layer because it offers the best steering trade-off between effect strength and performance degradation and corresponds to the mid-to-late layers that are commonly found most informative in probing studies.

\subsection{Probe Types and Regularization}

\begin{figure*}
    \centering
    \includegraphics[width=0.9\linewidth]{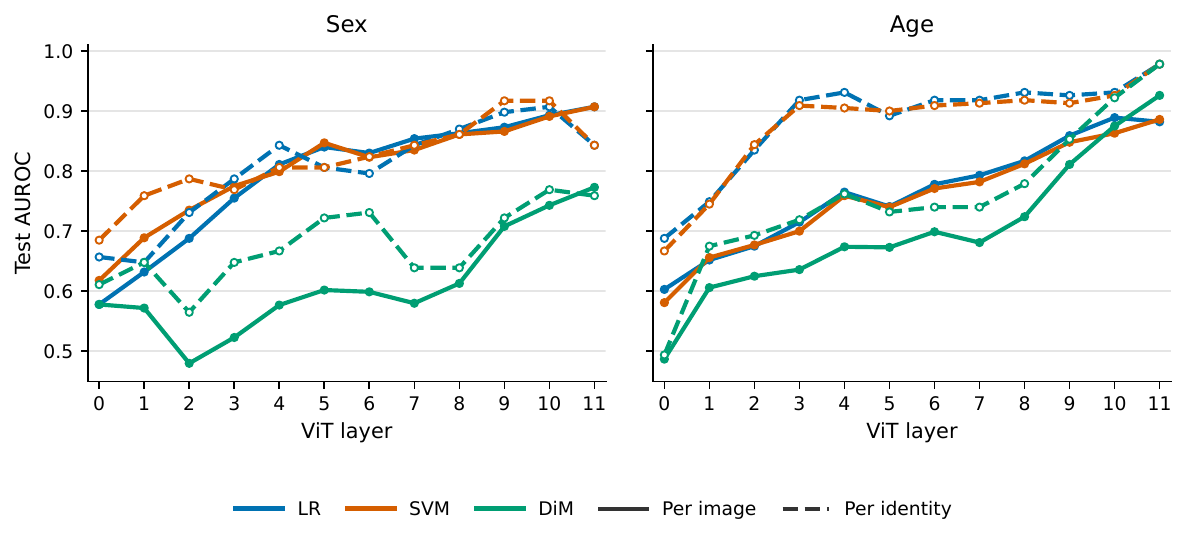}
    \caption{Layer-wise linear  separability of the sex and age concepts for the three probe types}
    \label{fig:probe_types}
\end{figure*}

We verify that the concept directions reported in the main paper do not depend on the particular probe used to extract them. Figure \ref{fig:probe_types} compares logistic regression (LR), linear SVM, and difference-in-means (DiM) probes across layers for both concepts, under both per-image and per-identity aggregation. The two optimized probes are effectively interchangeable: LR and SVM track each other closely at every layer, and their AUROCs coincide almost exactly across the second half of the network, where the concept is most separable. The learned directions are also near-collinear in these layers, so the choice between LR and SVM has no material effect on either separability or the direction we go on to steer along and attribute. We therefore report LR throughout the main paper and provide the SVM curves here for reference.

The parameter-free DiM direction behaves differently, recovering the concept substantially less well and peaking at only 0.77 AUROC for sex. This indicates that the concepts are encoded linearly but are not simply the axis joining the two class means: recovering them requires an optimized decision boundary rather than a mean-difference heuristic.

\begin{figure}
    \centering
    \includegraphics[width=0.9\linewidth]{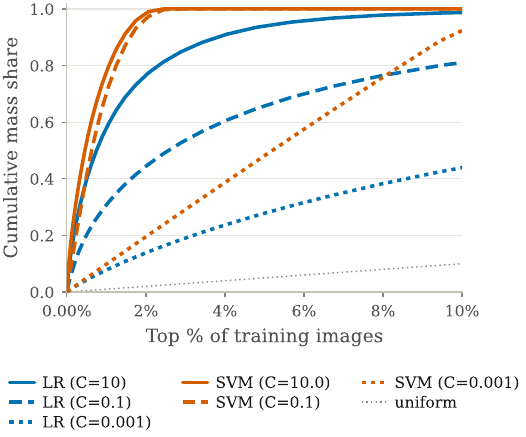}
    \caption{Cumulative sum of attributions against x\% of training images included at layer 7. Dashed lines indicate regularization C (0.001, 0.1 and 10), colors indicate probe type.}
    \label{fig:cumulative_sum}
\end{figure}

Separability is likewise insensitive to the strength of the $\ell_2$ regularization (Figure \ref{fig:cumulative_sum}). Sweeping the regularization parameter changes held-out AUROC by only 0.008 on average across the second half of the network. A visible effect appears only in the earliest layers, where stronger regularization underfits the (already weak) representation and lowers AUROC. Regularization does, however, change \textit{which} training images constitute the probe.

\subsection{Pretrained versus Fine-Tuned Representations}

\begin{figure}[t]
\centering
\includegraphics[width=0.85\linewidth]{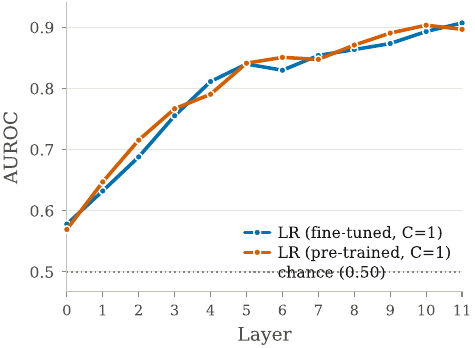}
\caption{Comparison between an LR probe fitted on pre-trained-only activations vs on fine-tuned activations.}
\label{fig:auroc_comp_ft}
\end{figure}

\begin{figure}[t]
\centering
\includegraphics[width=0.9\linewidth]{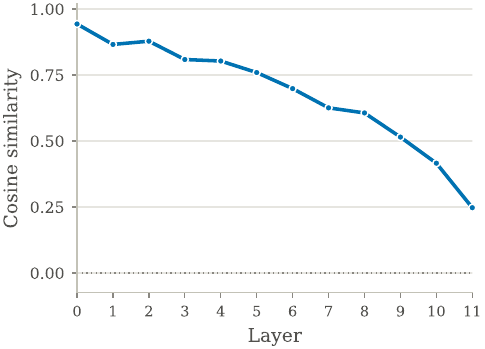}
\caption{Cosine similarity between concept directions learned on pretrained and Re-ID-fine-tuned activations.}
\label{fig:app_ft_cosine}
\end{figure}

We found that the probe fitted on the off-the-shelf DINOv3-Base (no fine-tuning) achieves similar performance in AUROC (Fig.~\ref{fig:auroc_comp_ft}).
We report the layer-wise divergence in cosine similarity between an LR probe trained on activations from the fine-tuned and the stock pre-trained DINOv3-Base model in Figure~\ref{fig:app_ft_cosine}.

\subsection{Sensitivity of Activation Steering}

\begin{figure*}
    \centering
    \includegraphics[width=0.9\linewidth]{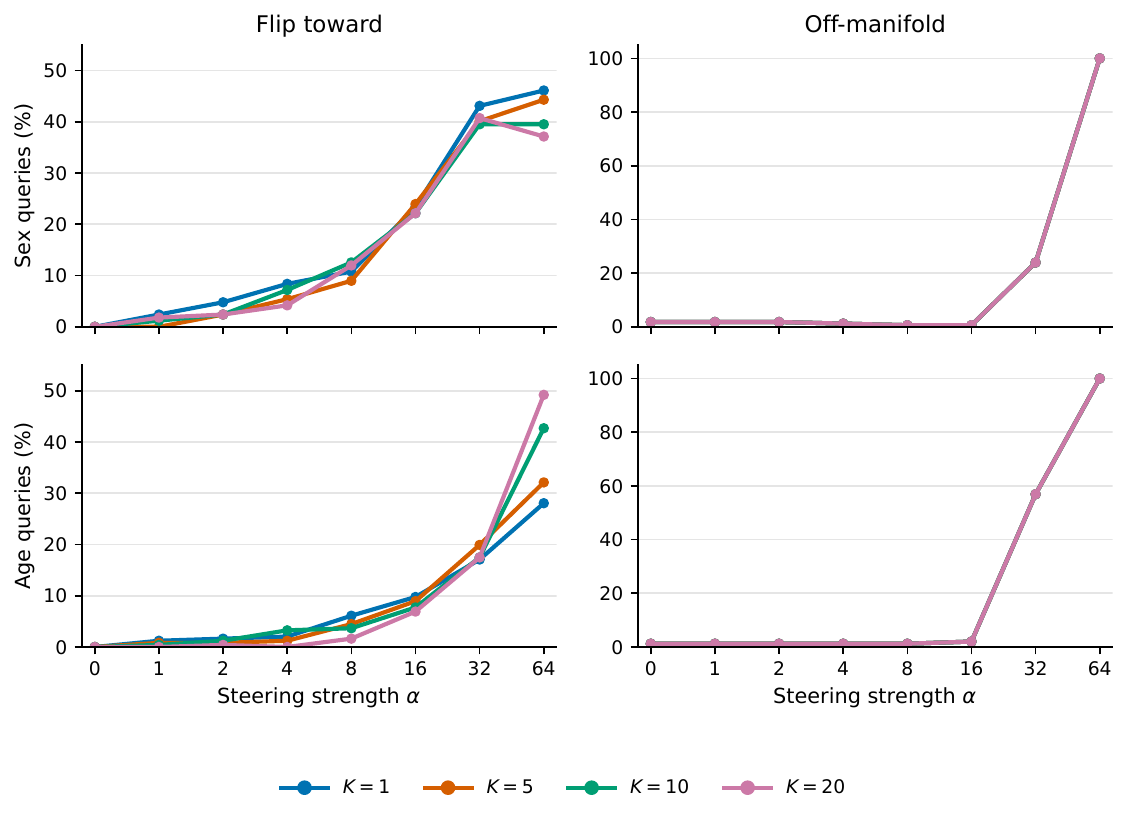}
    \caption{Steering results for LR sex and age probes at layer 9 across different values of $K$ and steering strength $\alpha$. All results are reported as percentages.}
    \label{fig:steering}
\end{figure*}

We report steering results for both concept LR probes at our standard steering layer 9, for multiple $K$ values in Figure \ref{fig:steering}. Steering result is stable with respect to the retrieval neighborhood size $K$. For both concepts, the flip-toward curves for $K \in \{1, 5, 10, 20\}$ exhibit a similar upward trajectory as the steering dose $\alpha$ increases and remain close to one another at every dose, while the off-manifold degradation curves are effectively indistinguishable across $K$, all turning upward only at $\alpha = 32$ and saturating near 100\% at $\alpha = 64$. The $K = 5$ value used in the main experiments is therefore not a special choice: neither the onset of a meaningful steering effect nor the point at which the representation collapses depends on how many neighbors the KNN prediction aggregates. The comparison also reproduces the concept-level asymmetry discussed in the main paper: the age direction needs a larger steering dose to move the same fraction of neighbors and only reaches an effect once degradation is already substantial, whereas the sex direction produces a targeted effect well before collapse. This asymmetry holds uniformly across all values of $K$.

\subsection{Robustness of Attribution-Set Composition}

\begin{figure*}
    \centering
    \includegraphics[width=0.9\linewidth]{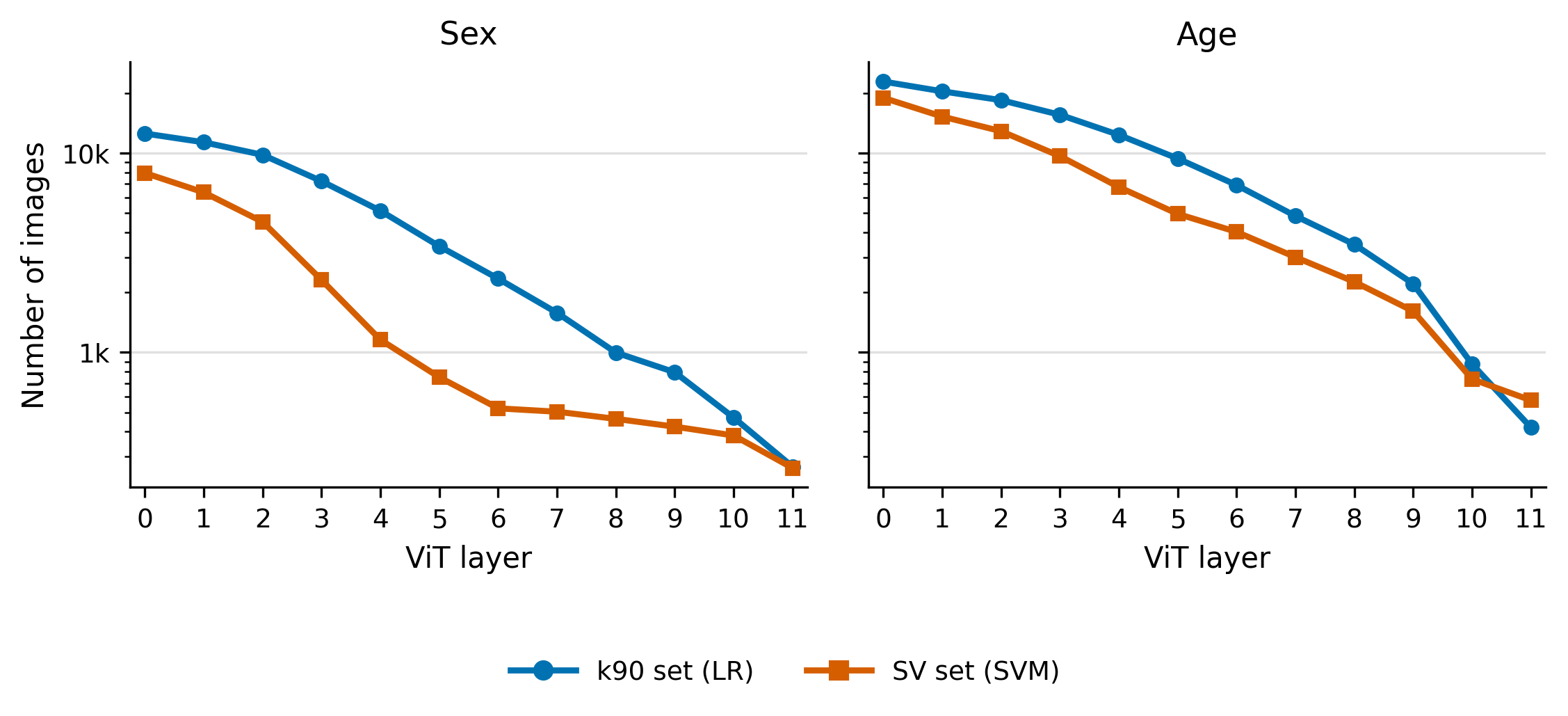}
    \caption{Number of images in the SV vs k90 sets for sex and age concepts}
    \label{fig:sv_vs_k90}
\end{figure*}

In our results, we used the set of support vectors (SVM probes) and our defined k90 set (LR probes) as largely interchangeable. Figure~\ref{fig:sv_vs_k90} shows that the numbers of images in each set are similar enough to make that generalization. Qualitative comparisons also showed that both approaches yielded similar results. 

\subsection{Individual-Level Composition}

\begin{figure}
    \centering
    \includegraphics[width=0.9\linewidth]{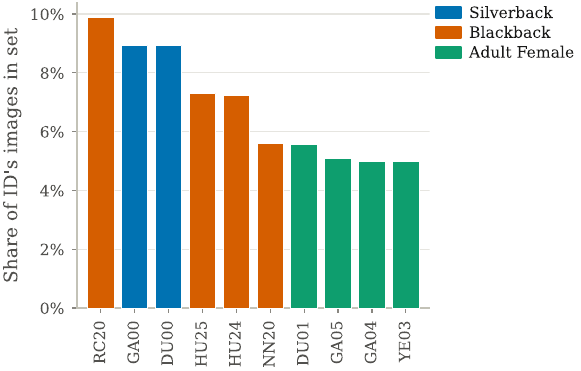}
    \caption{Percentage of each ID's total images selected to be support vectors (layer 9).}
    \label{fig:selection_rate_bars}
\end{figure}

We find that some individuals are overrepresented in the highest-attributing images.
Figure~\ref{fig:selection_rate_bars} shows the decomposition as the top 10 individuals with the highest selection rate. After normalizing for the unequal number of images per gorilla, the most frequently selected individuals are enriched by only 2 to 3 times relative to the uniform baseline, far from the near-exclusive concentration expected if the probes relied on memorized identities. Moreover, the LR (k90) and SVM support vector sets identify the same top ten individuals, indicating that the modest enrichment reflects shared representational structure rather than a method-specific selection bias.

\section{Implementation Details: DualXDA Attribution}
\label{app:implementation}

We use DualXDA \cite{yolcu2025sparseefficientexplainabledata} to propagate relevance scores back to the input images. Whereas the original XDA experiments employ convolutional backbones and the \texttt{zennit} library, our DINOv3 Vision Transformer requires a transformer-specific propagation scheme. We therefore apply AttnLRP \cite{achtibat2024attnlrp} using the \texttt{lxt} library.

Within each attention block, we adopt CP-LRP \cite{ali2022xai}, which treats the queries and keys as constants and prevents relevance propagation through the softmax operation. This configuration is provided by \texttt{lxt} for Vision Transformers and yields substantially better relevance conservation on our backbone than the full uniform rule. We apply the identity rule to LayerNorm and GELU, while leaving LayerScale unmodified because one of its operands is a learned parameter rather than an input-dependent activation.

Activations are extracted from unnormalized images with pixel values in $[0,1]$. A standard pixel-level Input$\times$Gradient attribution uses black as its reference point, assigning zero relevance to black pixels. This produces uninformative maps for our data, in which dark gorillas frequently appear against brighter backgrounds. We therefore use each image's mean color as the reference point for the pixel-level relevance expansion.

We note a limitation of DualXDA, which in our setting arises from the geometry of the learned representations. The heatmaps in the Test row in Figure~\ref{fig:dualxda_infant} are often highly similar across images, with their sign entirely determined by the concept label. This follows from the relevance initialization, which depends on the inner product $f_\text{test}f_i^\top$. Since the extracted representations have high pairwise cosine similarity, different support vectors $f_i$ and $f_j$ are approximately aligned. LRP propagates a seed proportional to $\lambda_i f_i$ for each support vector; hence, these aligned representations produce similar seeds and, consequently, similar relevance maps. The limited diversity of the Test heatmaps is therefore a consequence of the representation geometry rather than independent evidence from the individual support vectors.

\end{document}